\documentclass{article} 
\usepackage[final]{colm2026_conference}

\usepackage{microtype}
\usepackage{hyperref}
\usepackage{url}
\usepackage{booktabs}
\usepackage{graphicx}
\usepackage{amsmath}
\usepackage{wrapfig}
\usepackage{caption}
\usepackage{subcaption}

\usepackage{latexsym}
\usepackage{xcolor}
\usepackage{float}
\usepackage{enumitem}
\usepackage[LGR,T1]{fontenc}
\usepackage[greek,english]{babel}

\usepackage{lineno}

\definecolor{darkblue}{rgb}{0, 0, 0.5}
\hypersetup{colorlinks=true, citecolor=darkblue, linkcolor=darkblue, urlcolor=darkblue}

\title{An Investigation of Translationese in the Generations of \\ Multilingual Large Language Models}

\author{
Maria Valentini$^{1}$, Téa Wright$^{2}$, Julisa Granados$^{1}$, Eliana Colunga$^{1}$
\\ \& \textbf{Katharina von der Wense}$^{1,3}$
\\
$^{1}$University of Colorado Boulder \\
$^{2}$University of California Berkeley \\
$^{3}$Johannes Gutenberg University Mainz \\
\texttt{\{first.last\}@colorado.edu}
}

\begin{document}

\ifcolmsubmission
\linenumbers
\fi

\maketitle

\begin{abstract}
Text which has been translated from another language tends to carry with it evidence of translation — hence, it is often referred to as \textit{translationese}. Multilingual large language models (MLLMs) generate text in a variety of languages. However, it is still unclear if MLLMs' generations resemble internal translation (from English or, potentially, other languages) and, thus, result in translationese. Here, we ask the following research questions: (1) Does text generated by MLLMs resemble translationese? (2) How does translationese produced by MLLMs differ from translationese produced through direct translation? We leverage established indicators of translated text to evaluate text generated by state-of-the-art MLLMs in five languages, comparing to both non-translated and human-written baselines in order to isolate translationese from other kinds of interference. Through the use of high-accuracy classification models, analyses of variance on individual linguistic features, and the collection of human annotations for a subset of two languages (German and Spanish), we assess the translationese content of MLLM generations and examine the key features that distinguish MLLM-generated text from typical translation-related interference.
\end{abstract}

\section{Introduction}

It is well documented that large language models (LLMs) are becoming increasingly adept at generating text in non-English languages. Systems such as ChatGPT \citep{chatgpt}, while not explicitly designed for multilingual generation, have demonstrated some level of ability to generate text in non-English languages. Other models such as PaLM2 \citep{palm2} are explicitly designed for multilingual use and claim improved fluency for many languages when compared to their counterparts which are primarily trained on English. 

The linguistic naturalness of LLM-generated text is an important component of its overall quality, highly relevant for human-facing applications such as designing language learning materials. However, it is often overshadowed by more traditionally important attributes, such as correctness, coherence, and relevance. While these are all critical for assessment, LLMs should ideally also mirror the naturalness of native speakers, regardless of language.

In recent years, apparent multilingual proficiency in MLLMs has raised the question of whether they truly generate text in non-English languages in a linguistically native way or whether their outputs instead reflect translation from a dominant internal language (e.g., English). Given the black-box nature of LLMs, it is difficult to know exactly what is going on behind the scenes when text is generated. However, recent work suggests that MLLMs may rely on English-centric internal representations, even when prompted in another language \citep{schut2025multilingual}. 

\begin{wrapfigure}[14]{r}{0.5\textwidth}
\centering
\vspace*{-1.2cm}
\captionsetup{width=.9\linewidth}
\includegraphics[width=0.48\columnwidth,height=0.8\columnwidth,keepaspectratio]{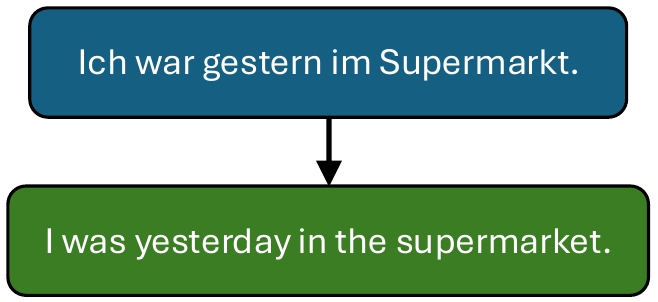}
    \caption{An example of translationese interference. This example shows a German-to-English translation with interference in part-of-speech n-gram distribution, which can be automatically detected with various machine learning models.}
    \label{translation}
\end{wrapfigure}

Here, we take a different approach and focus on MLLM \emph{generations}: we evaluate LLMs using externally observable phenomena typical for translated text, known as translationese. This has potential impact both in assessing LLM naturalness for human-facing applications and interpretability research. Fig. \ref{translation} shows a simple example of translationese.

In this paper, we present several experiments, with the goal of answering the following research questions: (1) Does text generated by MLLMs resemble translationese? (2) How does translationese produced by MLLMs differ from translationese produced through direct translation? In our first experiment, we focus on the high- and medium-resource languages of English, German, Spanish, and Greek. We compare LLM-generated text to multiple reference conditions, allowing us to examine whether translationese emerges in multilingual generation even for languages with substantial training data. We extend this analysis to a low-resource language, Pashto, to examine whether effects are exaggerated when there is less available pretraining data. We demonstrate that LLM-generated texts exhibit some level of translationese in all languages (even English), but its presence is particularly elevated in non-English languages. We further analyze individual linguistic features and explore structural differences across languages, positing explanations for variation and establishing methodologies for future work on evaluating LLM outputs for translationese.

\section{Related work}

\paragraph{Translationese} Translationese refers to the 'fingerprint' that remains after text has been translated \citep{Gellerstam}. It is suggested to be a result of adhering to the meaning of the source text while adapting to the grammatical structures of the target language \citep{Toury1980}. This includes both general effects of the process of translation independent of source language and the ways in which a specific source language leaves distinct marks in the target language, also known as \textit{interference} \citep{Koppel, Toury_1995}. \citet{Volansky} demonstrates that there are distinct features of translationese that can be indexed by linguistic indicators relating to principles such as simplification and explicitation, both of which are thought of as universal for translation. Even text which has been translated by a machine and reviewed/edited by a human has still been shown to demonstrate translationese \citep{Toral_2019}.

\paragraph{Automatic Detection of Translationese}

Several works have examined how to use computational methods to automatically detect or measure translationese. \citet{Baroni} and \citet{Freitag_2022} utilize a support vector machine which distinguishes between translated text and its original with high accuracy. \citet{Volansky} trains classifiers on the linguistic features previously identified to be hallmarks of translation, as mentioned above. These include indicators such as type--token ratio as well as sentence and word length based on the simplification hypothesis \citep{baker1993} and indicators such as POS n-gram distribution based on interference \citep{Toury_1995}. This methodology for translationese identification is also adapted by \citet{Hu_2021} and \citet{Rabinovich_2017}. \citet{Borah_2023} supports the use of these surface features, establishing that semantic features can introduce spurious topic-based biases. Finally, recent studies such as \citet{Riley2020} and \citet{Jalota} demonstrate further that translationese can be effectively quantified, even without a parallel corpus.

\paragraph{Multilinguality and Large Language Models}

Prior work shows that MLLMs struggle with capturing cultural nuances and balancing universal knowledge with language-specific details, especially for low-resource languages --- challenges directly relevant to translationese, given its ties to linguistic fidelity \citep{li2024culturellm, held2023material}.
Another recent line of work on MLLMs investigates how multilingual capabilities may be related to translation, both internally and externally. For example, \citet{etxaniz-etal-2024-multilingual} observes that inference strategies such as translating inputs into English improves performance in multilingual settings. \citet{Li_Zhang_Wang_Zhang_Cui_Yin_Xiao_Zhang_2025} similarly finds that LLM translation exhibits translationese, tracing it to biases from supervised fine-tuning rather than inference-time processing alone. Another study shows that MLLMs pass tokens through English representation space even when prompted in other languages. This effect becomes more prominent the less diverse pretraining the model had \citep{schut2025multilingual}. Together, these findings indicate that multilingual generation may involve English-centric processing or implicit translations.

\section{Human-written Data}
\label{data}

\subsection{German, Spanish, and Greek Data} 
For German, Spanish, and Greek, we leverage both native-speaker and non-native, human-translated text from the Europarl dataset \citep{koehn-2005-europarl}, an extensive multilingual parallel corpus collected from proceedings of the European parliament from 1996 to 2012. We tokenize the text and partition it into chunks of approximately 2000 tokens (ending on a sentence boundary) to ensure that the length of an article does not interfere with classification. 

The resulting German and Spanish datasets each contain 8000 samples, including 4000 natively written ones and 4000 ones that have been translated into German or Spanish from English. We use an 80/20 train/test split, resulting in 6400 and 1600 instances for training and testing, respectively.\footnote{We use 10-fold cross-validation for model development.}

In our test sets, we also include a subset which extracts the non-native, human-written text \textit{before} it has been translated into the target language, translating it instead using Google Translate. Including this condition is necessary to disentangle translationese effects from LLM-generation effects; it allows us to test whether the classifier's signal is driven by:

\begin{enumerate}[label=(\alph*),leftmargin=*, noitemsep, topsep=0pt, partopsep=1pt]
    \item \textit{Translationese}, which should affect both human + MT and LLM + MT texts similarly, or
    \item \textit{LLM-specific stylistic biases}, which would cause the LLM + MT condition to diverge from the human + MT baseline even though both underwent identical MT processing.
\end{enumerate}

Finally, we include configurations where the native human text is translated by either Gemini or Llama, in order to examine any potential model-specific translation biases.

For Greek, while the data is also processed in the way described above, we only obtain a  dataset of size 6000 (with a 4800/1200 train/test split). Though we obtain data from the same source as for German and Spanish, we note that on Microsoft's linguistic diversity index \citep{ling_diverse}, in which 5 is the class with most resources and 0 is the one with the fewest, Greek is classed at 3, meaning it has somewhat limited resources, but still more than nearly 95\% of the world's languages.

\subsection{English Data} 
We also develop an English dataset for comparison, to ensure that interference we measure is due to translationese as opposed to interference purely from LLM generation, which would show up in any language. We use Europarl for this as well, collecting a dataset matching the proportions of our German dataset (6400/1600). 4000 of these are original English writing, while 4000 are translated from the ten languages used to train the \citet{Volansky} classifier (400 each). 

\subsection{Pashto Data} 
We conduct an additional experiment in which we focus on the low-resource language Pashto, an eastern Iranian language with around 45-55 million speakers worldwide. In Microsoft's linguistic diversity index \citep{ling_diverse}, Pashto is classified at 1 on the 0-5 scale, meaning their digital resources are very limited in comparison to English, Spanish, German, and even Greek.

For the Pashto language, we collect native speaker data primarily by scraping Pashto  Wikipedia articles. We begin on a popular page and randomly select from the links it includes, then travel to that page and repeat the process. We extract the first 10 sentences corresponding to each article title gained through the random crawl. If an article consists of fewer than 10 sentences, we reject it and move to a different one. In total, we collect samples from 500 articles.

To obtain data which has been human-translated into Pashto, we utilize the Open Language Data Initiative (OLDI) Seed dataset \citep{seed-23}. The OLDI Seed dataset consists of roughly 6,193 sentences drawn from English-language Wikipedia and translated into 44 diverse languages. As the entries in the dataset are stored at the sentence level, we recombine them by article and split them into 10-sentence-long chunks. This results in a total of 500 human-translated Pashto samples from English Wikipedia articles. Our final Pashto human dataset consists of 1000 samples, with a 800/200 train/test split, where the split is partitioned at the article level so that no chunks from the same source article appear in both sets. While not directly parallel like our other datasets, comparable corpora such as this one have demonstrated effectiveness for translationese detection \citep{Riley2020, Jalota}.

\subsection{Europarl-UdS}

Finally, we include an additional experiment to further verify our Spanish and German results in Appendix \ref{sec:uds}. This experiment utilizes the Europarl-UdS dataset \citep{uds}, a specifically curated version of Europarl which isolates texts confirmed to be uttered by a native speaker. We cannot use UdS to replace our primary evaluation source as it is currently only available for English, German, and Spanish, and only provides English-to-target-translated parallel data, making it insufficient for our complete multilingual evaluation (our English verification baseline also cannot be fully replaced by UdS, with English only on the source side of the parallel data).

\section{Experiments}
Our goal is to assess the levels of translationese in MLLM generations for several languages with varying levels of language-specific training text available.
We utilize samples in each language generated by the following configurations: (1) MLLM generating in target language, (2) MLLM generating in English + MT system translator, (3) target-language native speaker, (4) English native speaker + human translator, (5) English native speaker + MT system translator, (6a) English native speaker + Gemini translator, and (6b) English native speaker + Llama translator.

We attempt to select a diverse set of languages in order to ensure observations are valid across different languages; those selected differ with regards to their linguistic diversity index, their morphosyntactic complexity, and their similarity to English. We include configurations with both human and machine translation in order to ensure the classifier can accurately classify both, as well as any LLM-specific differences that may arise.

\subsection{Measuring Levels of Translationese}

This section outlines the metrics we use to evaluate text generated by all models or model combinations for the presence of translationese, which include individual linguistic indicators as well as classification results from a high-accuracy binary classifier.

\subsubsection{Linguistic Indicators}
\label{indicators}
For all experiments, we use indicators which have been selected for their relevance to the process of translation, and narrowed down further by us based on the accuracy scores reported by \citet{Volansky}. The use of these surface features rather than deeper semantic features allows us to avoid potential topic-based biases. In addition to using these indicators as features for our classifier, we conduct analyses of variance (ANOVAs) on individual categories to extract more fine-grained information on their specific relevance. The indicators we implement are described below, as well as their reported accuracies.

\begin{itemize}[leftmargin=25pt]
    \item \textbf{Type-token ratio}: ratio of types (unique tokens) to total tokens, measuring lexical variety (\citet{Volansky}: $72\%$ accuracy).
    \item \textbf{Function words}: ratio of function (non-content) word tokens to total tokens, identified via POS-tagging per \citet{Baxronovish_2016}. Each unique function word is treated as its own classifier feature, as in \citet{Volansky} ($96\%$ accuracy).
    \item \textbf{Mean sentence length} ($65\%$) and \textbf{mean word length} ($66\%$): average lengths of sentences and words per passage.
    \item \textbf{Pronoun frequency}: ratio of pronouns to total tokens; we combine per-pronoun features (individually $77\%$ in \citet{Volansky}) into a single indicator.
    \item \textbf{Punctuation frequency}: ratio of punctuation marks to total tokens ($81\%$).
    \item \textbf{POS n-gram distribution}: correlation of unigram/bigram/trigram POS distributions with typical distributions for each language, split by unique n-gram as a classifier feature ($90\%$/$97\%$/$98\%$). We tag with StanfordPOSTagger (German, Spanish), spaCy 3.7.5 (Greek), and a BERT-based XLM-R tagger \citep{ijazul_2025} (Pashto).
    \item \textbf{Contextual function words}: triplets with at most one POS-tagged word and at least two function words, split by unique triplet as a feature ($100\%$).
\end{itemize}

\subsubsection{Translationese Classifier}

The primary method we use to evaluate each of our generation configurations is a binary support vector machine (SVM) classifier, which we train by implementing a set of established translationese indicators as features (see Section \ref{indicators}). More specifically, we implement John Platt's sequential minimal optimization (SMO) algorithm \citep{smo_platt}. To ensure that the classifier captures the general linguistic signal of translationese rather than artifacts of a specific MT system, we train the SVM exclusively on human-produced data and translations. This mitigates the risk of inducing a decision boundary that reflects idiosyncratic properties of a specific MT system---its lexical preferences, sentence-length distributions, or characteristic error modes---rather than structural properties associated with translated text more broadly.

We train a classifier for each language (German, Spanish, Greek, English, and Pashto) using their respective training sets, which are described in Section \ref{data}. We note the overall accuracy scores achieved by each language's classifier in Table \ref{accuracies}. This same information for the UdS dataset is provided in Appendix \ref{sec:uds_descrip}.

\begin{wraptable}[13]{l}{0.36\textwidth}
\centering
\captionsetup{width=.7\linewidth}
\begin{tabular}{lc}
\toprule
\textbf{Language} & \textbf{Accuracy (\%)} \\
\midrule
German  & 99.79 \\
Spanish & 99.81 \\
Greek   & 99.95 \\
Pashto  & 87.50 \\
English & 99.36 \\
\bottomrule
\end{tabular}
\caption{SVM classification accuracy by language, in percentages.}
\label{accuracies}
\end{wraptable}

Once a model is trained for each language, we are able to evaluate configurations present in the test set using the \textbf{Classified Positive} metric, which refers to the proportion of samples generated using each configuration which are labeled by the SVM as positive for translation. 
\vspace*{-.2cm}

\paragraph{Feature Coefficients} Furthermore, feature coefficients allow us to determine how influential each specific feature is in measuring the presence of translationese. When an SVM with a linear kernel is trained, it generates a boundary that specifies weights for each individual feature. The magnitude (absolute value) represents how strongly each normalized feature changes the score per unit increase, holding all other features fixed. A positive coefficient means that a feature pushes the classifier toward a \textit{True} classification (indicating translated text) and a negative coefficient means that a feature pushes the classifier toward \textit{False} (native text).

\subsection{Data Generation}
This section outlines the construction of the machine-generated portion of our datasets.

\subsubsection{Models}
\label{LMs}

\paragraph{Gemini} We use the Gemini-2.0-Flash-001 model \citep{Gemini} as one of our MLLMs  because of its strong multilingual claims. Gemini-2.0-Flash is part of the Gemini model family, purportedly more successful than previous models with multilingual generation due to the large amount of multilingual data included in the training corpus.
\vspace*{-.1cm}

\paragraph{Llama3} We additionally compare to Llama-3.3-70B-Instruct \citep{llama}, an open-source model with strong multilingual capabilities, to make sure our findings generalize.
\vspace*{-.1cm}

\paragraph{Generation+Translation Pipelines} We finally compare to a pipeline approach where both LLMs generate text in English and the Google Translate API translates the generations into the target language. We call these configurations \textit{Gemini+MT} and \textit{Llama+MT}.

\subsubsection{Process}

For each language, the set of LLM-generated texts corresponds directly to the human-written test set described in Section \ref{data}. Prompts are based on the content of each human sample in order to create corpora with as many similarities as possible, removing excess noise that might influence classification. This generation step marks the last stage of development for our dataset. The total number of samples now generated and collected for each language/configuration pair in the completed dataset can be found in Appendix \ref{sec:data_app}.

\subsubsection{Prompt Structure}

The English LLM prompt used to generate text to be translated into German, Spanish, and Greek was constructed as follows, with <\textit{first sentence}> representing the initial sentence of each sample in our pre-translation English Europarl test set:

\begin{itemize}[leftmargin=15pt]
    \item You are given the first sentence of a statement made in the European parliament. Write a new paragraph using the same style that could complete the statement. Continue your response until you hit 2000 tokens, then truncate to the last sentence's end. Return ONLY the paragraph with no additional text. The sentence to expand on is: <\textit{first sentence}>.
\end{itemize}

\noindent The LLM prompt used for English generation into Pashto was constructed as follows, with <\textit{Title}> representing the title of each pre-translation English article in our OLDI Seed test set:

\begin{itemize}[leftmargin=15pt]
    \item Write the first 10 sentences for a Wikipedia article with the title: <\textit{Title}>. Return ONLY the sentences with no additional text.
\end{itemize}

For direct generation in each target language, we translate these prompts using Google Translate. We verify these translations by back-translating into English for review. Finally, in German and Spanish, we asked native speakers to 'internalize,' or memorize the task itself (rather than the specific words used to describe it), and then write in their own words and in their native language a description of the task, in order to ensure the LLM outputs are not influenced by potential translationese within the prompts.

\subsection{Human Annotation Data}

We additionally create a native speaker-annotated subsection of our German and Spanish test sets. For each language, we randomly sample 60 sentences from the possible configurations (see Table \ref{datatable} for a full list of the configurations making up each set). Our annotators are given the sentences individually, with no additional context or labels, and instructed to label each sentence as either \textit{Translated} or \textit{Non-Translated}. If a sentence is labeled as \textit{Translated}, they are asked to provide a few words in a separate column explaining their reasoning. This annotated subset allows for a small analysis of how well humans are able to detect translationese, which sorts of features are more and less noticeable to human readers, and how LLM outputs compare with human translationese preferences. We note that all annotators are authors of this paper who are university-educated and native speakers of their respective languages.

\subsection{Statistical Analysis}

We conduct Analyses of Variance (ANOVAs) on the linguistic indicators described in Section \ref{indicators} in order to obtain more specific information on how each machine generation configuration differs from human text, both translated and non-translated. We measure the statistical strength of each indicator in differentiating translated from non-translated text, also examining how text generated using each LLM-based configuration compares in score to both human settings. 

This allows for a more fine-grained analysis of \textit{how much} and \textit{what kind of} translationese is present in text generated using each configuration. For each language, we examine ANOVA results of indicators which show particular strength in separating translated from non-translated text. Comparing how configurations score on these indicators allows us to isolate the features which are and are not displayed in LLM-generated translationese, addressing our second research question (\textit{How does translationese produced by MLLMs differ from translationese produced through direct translation?}).

\section{Results and Discussion}

\subsection{Classification Results}

\subsubsection{English}

\begin{wrapfigure}[31]{r}{0.6\textwidth}
\centering
\vspace*{-3.25cm}
\captionsetup{width=.9\linewidth}
\begin{minipage}{\linewidth}
    \centering
    \includegraphics[width=0.95\columnwidth,height=0.9\columnwidth,keepaspectratio]{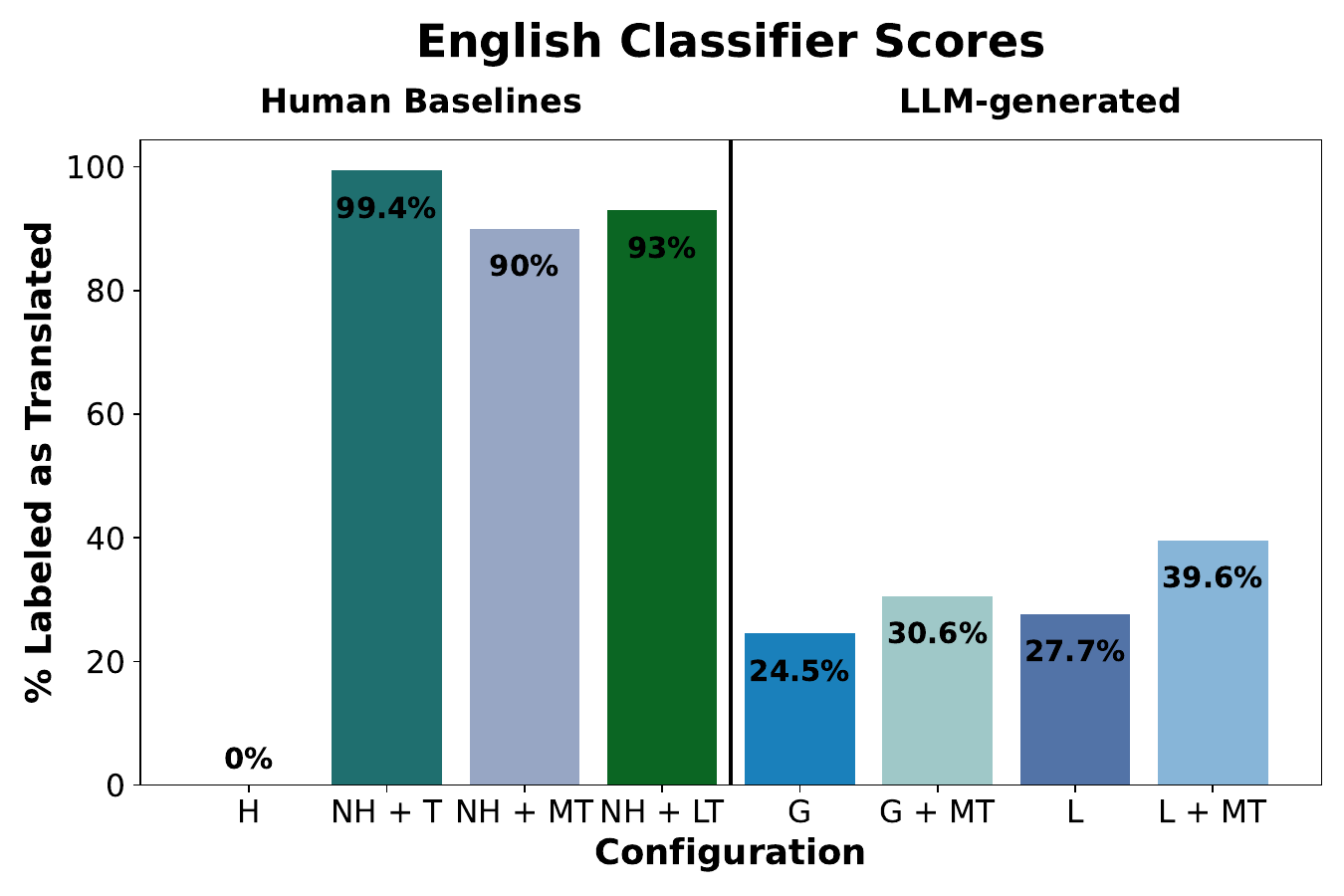}
    \vspace*{-.05cm}
    \caption{Full results for the English translationese classifier (higher = more samples classified as translated). \textit{H} = human, \textit{NH} = non-native human, \textit{G} = Gemini, \textit{L} = Llama, \textit{T} = human translation, \textit{MT} = machine translation, \textit{LT} = average of Llama and Gemini translation scores.}
    \label{english}
\end{minipage}
\begin{minipage}{\linewidth}
    \centering
    \vspace*{.4cm}
    \includegraphics[width=0.95\columnwidth,height=0.9\columnwidth,keepaspectratio]{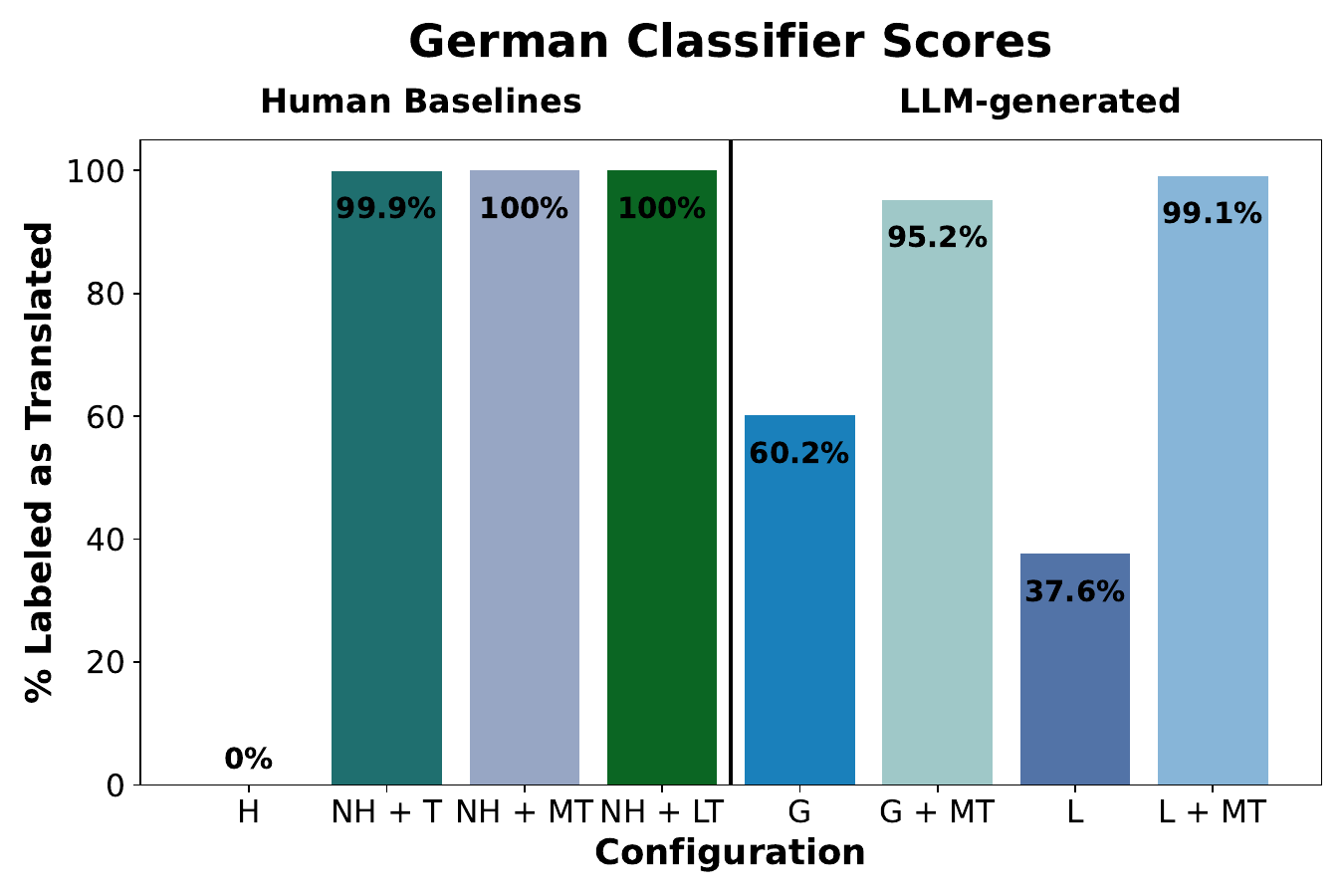}
    \vspace*{-.05cm}
    \caption{Full results for German. See Figure \ref{english} for the complete dictionary of abbreviations.}
    \label{german}
\end{minipage}
\end{wrapfigure}

Our English classification results (see Figure \ref{english}), demonstrate that LLMs may naturally produce some level of interference resembling translationese, even when generating English. 
However, our results on Europarl-UdS data suggest that this may also just be dataset-specific noise, as the classifier trained on the filtered data picked up little to no translationese in English LLM-generated texts (see Appendix \ref{sec:uds_results}).

The English results included here provide a comparison for our multilingual results, allowing us to disentangle the base effects of LLM generation from non-English generations.

We additionally note the effectiveness of the classifier at properly classifying each of our human baselines. Despite not having encountered any machine-translated text during training, it is still able to identify it as translated with $90\%$ accuracy. This indicates the classifier does in fact appear to be picking up on artifacts of translation, not just idiosyncrasies of the dataset or individual translators. The LLM-translated results are similar (averaging within 8 percentage points for all languages in our primary dataset), and are included in full in Appendix \ref{sec:llm_trans}.

\vspace*{-.05cm}
\subsubsection{German}

For German (see Figure \ref{german}), translationese levels are considerably higher in both MLLMs, with Gemini and Llama having $60.2\%$ and $37.6\%$ of their outputs scored by the classifier as translated, respectively. This suggests MLLMs may inherently struggle more with non-English naturalness, despite multilingual training. This same finding is reflected in UdS experiments, with German generations scoring as translated $18$-$26\%$ more than in English. 

Several factors could explain this pattern. One hypothesis, which has also been studied recently in various interpretability research (e.g., \cite{wendler2024llamasworkenglishlatent} or \cite{schut2025multilingual}), is that MLLMs may actually be 'thinking' in English, performing the majority of reasoning steps in English before translating into the desired generation language. This would likely yield an increase in translationese scores when generating in non-English languages, similar to what we see here for German. Given these results in complement with existing research, this seems like a plausible hypothesis.

\begin{wrapfigure}[16]{r}{0.6\textwidth}
\centering
\vspace*{-.5cm}
\captionsetup{width=.9\linewidth}
\includegraphics[width=0.57\columnwidth,height=0.9\columnwidth,keepaspectratio]{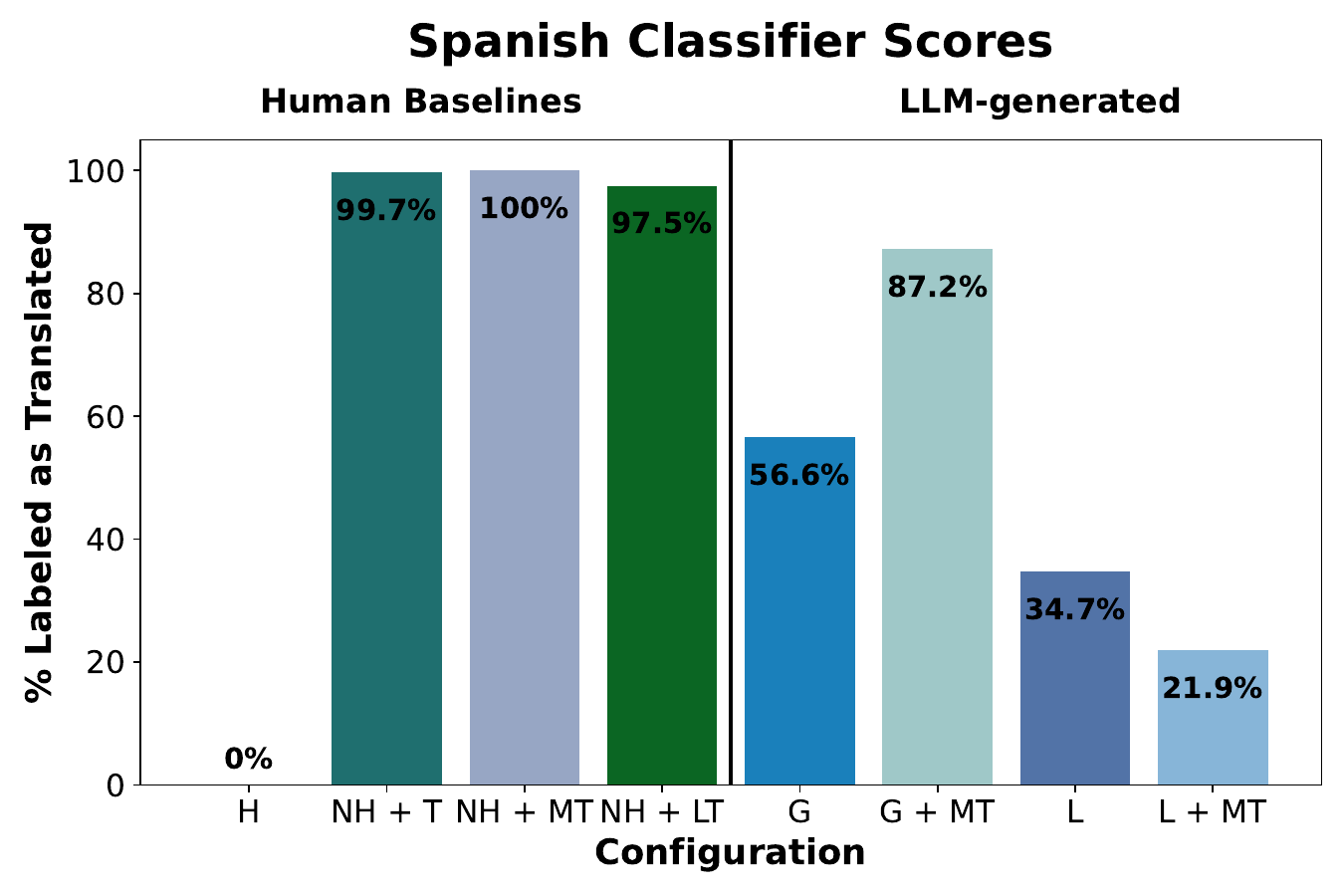}
    \caption{Full results for Spanish. See Figure \ref{english} for the complete dictionary of abbreviations.}
    \label{spanish}
\end{wrapfigure}

Another explanation, however, could just be that the data these models are trained on contains large amounts of translated text, and the text they generate is only mimicking behavior from their training data. It is thus important to acknowledge that these results alone do not necessarily indicate text has been translated, but they represent a linguistically grounded evaluation that can provide support to existing theories.

\subsubsection{Spanish}

Spanish results (see Figure \ref{spanish}), which yield an average score of $45.7\%$ translated across the two native LLM configurations, also suggest a natural inclination for LLMs to produce more translationese in non-English languages. These results are relatively similar to German, with the exception that the Llama-generated and 
machine-translated configuration outperformed its non-translated counterpart. This is unexpected, but a closer analysis of the text generated by this configuration reveals that article usage is particularly low, which may be a Llama-specific peculiarity. We expand more on this idea in Section \ref{anova}. As with German, these results are reflected in the filtered UdS data, including the Llama+MT abnormality.

\subsubsection{Greek and Pashto}

\begin{wrapfigure}[14]{r}{0.6\textwidth}
\centering
\vspace*{-1.5cm}
\captionsetup{width=.9\linewidth}
\includegraphics[width=0.57\columnwidth,height=0.9\columnwidth,keepaspectratio]{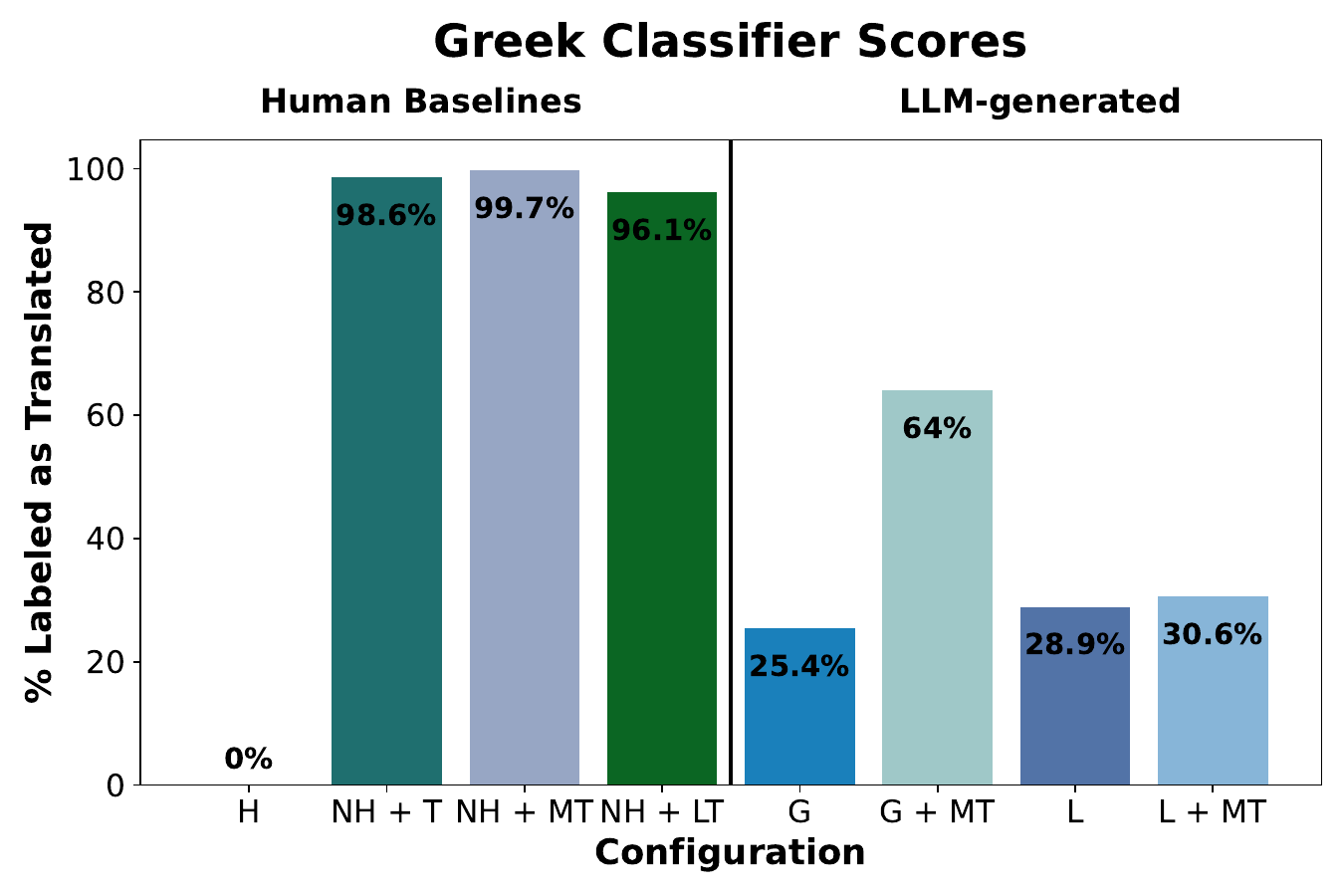}
    \caption{Full results for Greek. See Figure \ref{english} for the complete dictionary of abbreviations.}
    \label{greek}
\end{wrapfigure}

\begin{wrapfigure}[17]{r}{0.6\textwidth}
\centering
\vspace*{-.3cm}
\captionsetup{width=.9\linewidth}
\includegraphics[width=0.57\columnwidth,height=0.9\columnwidth,keepaspectratio]{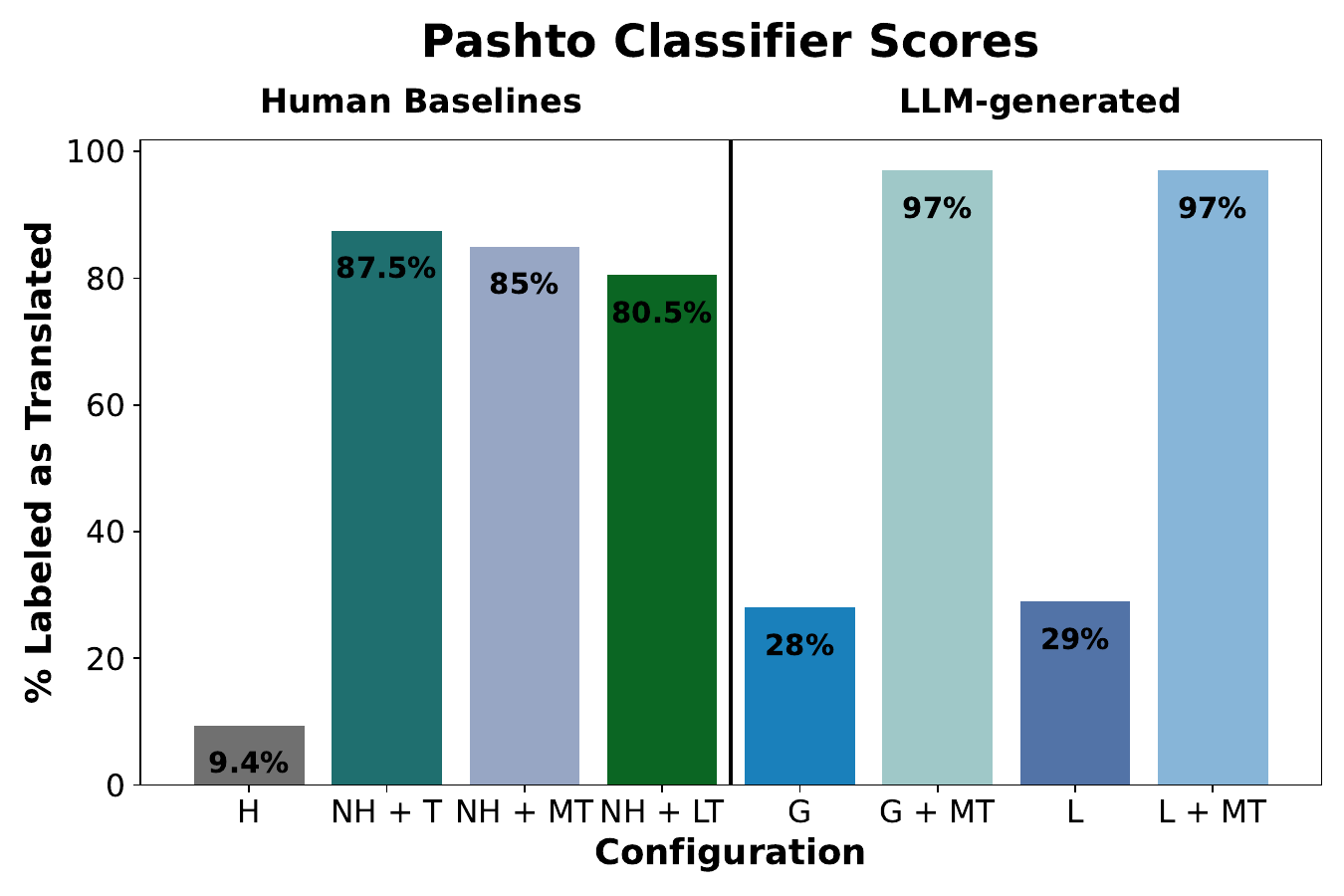}
    \caption{Full results for Pashto. See Figure \ref{english} for the complete dictionary of abbreviations.}
    \label{pashto}
\end{wrapfigure}

In both Greek and Pashto (see Figures \ref{greek} and \ref{pashto}), we observe similar results:  MLLMs are actually more successful at producing translationese-free text in Greek and Pashto than in Spanish or German, despite the latter two languages having considerably larger sets of data and computational resources available, as well as closer similarity to English. It is difficult to say exactly why this is, but there are a few aspects of these languages that could contribute.

For one, English--Spanish and English--German are common parallel-data pairings, which benefits NLP tooling but likely also means more translated text in MLLM training data. In languages with smaller amounts of resources, like Greek and Pashto, although both data quantity and quality may be degraded, the lack of translated data may be beneficial purely from a translationese generation standpoint. 

Additionally, Greek and Pashto are both languages with extremely flexible word orders \citep{WordOrderFlexibilityandAdjacencyPreferencesCompetingForcesandTensionintheGreekVP, Shah_2025}. While German also demonstrates flexibility with word order, it has characteristics that still require structured learning when transferring from English, like the V2 constraint and verb-final subordinate clauses.

Finally, there is a possibility that their greater similarity to English is precisely what causes more interference in German and Spanish; e.g., overlapping representational space may make English patterns harder for the model to suppress.

\subsection{ANOVA Results}
\label{anova}

Looking at features with the highest absolute weight, we see that trends do appear in the broader distribution/ratio-based indicators, but the most deterministic features are those which focus on something more specific, like the frequency of individual function words. In German, a higher frequency of the function word \textit{auch} is one of the strongest indicators that text has not been translated, with an attribute weight of $-0.327$ (the highest magnitude of any feature), while the function word \textit{äu{\ss}erst} is a stronger indicator of translated text, with a weight of $+0.200$.

\begin{wrapfigure}[17]{r}{0.6\textwidth}
\centering
\vspace*{-.6cm}
\captionsetup{width=.9\linewidth}
\includegraphics[width=0.57\columnwidth,height=0.9\columnwidth,keepaspectratio]{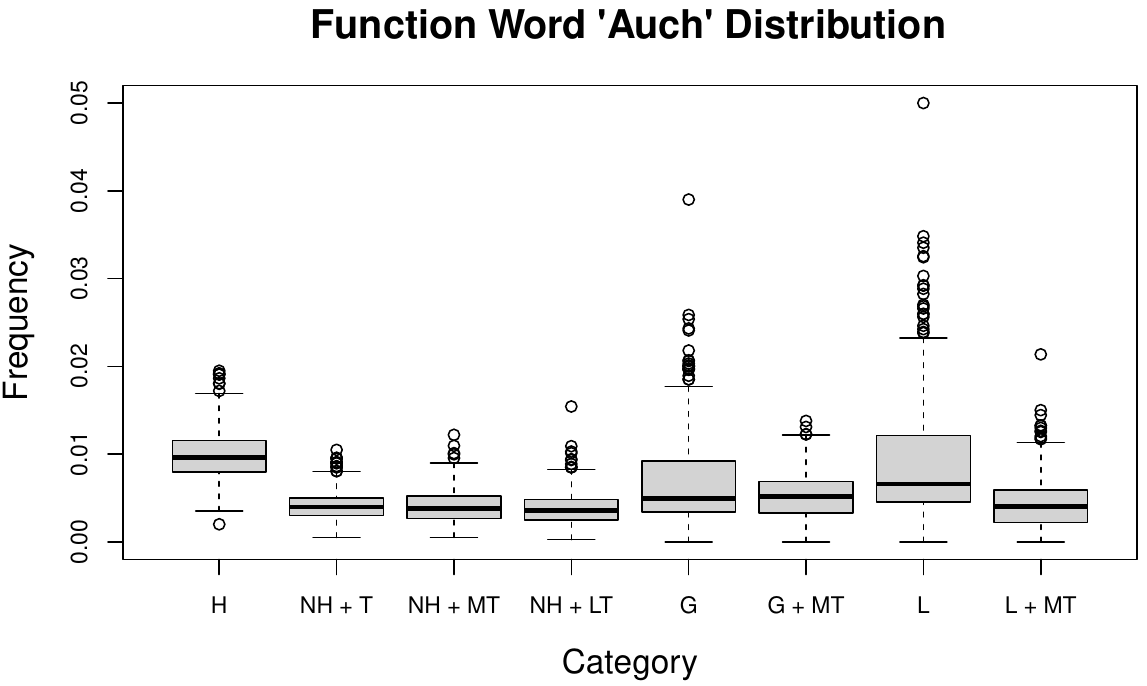}
    \caption{Distribution of the German function word \textit{auch}. See Figure \ref{english} for the complete dictionary of abbreviations.}
    \label{fig_auch}
\end{wrapfigure}

\begin{wrapfigure}[15]{r}{0.6\textwidth}
\vspace*{-.2cm}
\centering
\captionsetup{width=.9\linewidth}
\includegraphics[width=0.57\columnwidth,height=0.9\columnwidth,keepaspectratio]{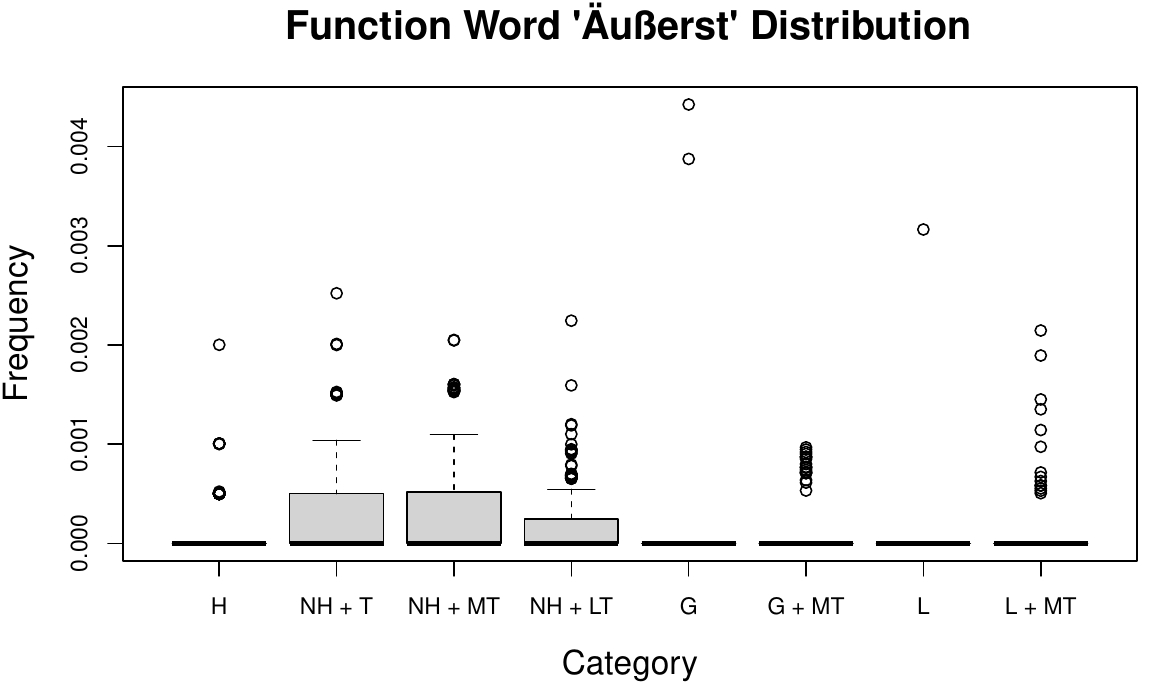}
    \caption{Distribution of the German function word \textit{äußerst}. See Figure \ref{english} for the complete dictionary of abbreviations.}
    \label{fig_aussrest}
\end{wrapfigure}

Comparing LLM-generated text to translated and non-translated text on these features reveals another trend. The models are generally able to avoid including features whose presence is a strong indicator of translation (as in the above example); the ways in which their generated text diverges from native human text are typically in the specific distribution of higher-frequency features. This is consistent with human-preference tuning in LLM training pipelines \citep{jiang2024surveyhumanpreferencelearning}, which likely filters out the most obviously abnormal features. This pattern is demonstrated in Figures \ref{fig_auch} and \ref{fig_aussrest}: LLMs are largely able to avoid infrequent and abnormal function words like \textit{äu{\ss}erst}, but the distribution shifts back toward matching translated texts for higher frequency words like \textit{auch}.

Spanish ANOVA results confirm that the number of articles (e.g., \textit{el}/\textit{la}/\textit{los}/\textit{las}) are indeed an abnormality in Llama configurations: the function word \textit{el}, for example, which has a feature weight of $+0.1192$ (pushing toward translated), has an aggregate average for Llama configurations equal to less than half that of Gemini or human outputs. For the full set of Spanish article ANOVA results, see Appendix \ref{sec:spanish}.

\subsection{Human Preferences}

Human annotations suggest that some LLMs may indeed largely filter out translationese cues more obvious to humans: while $60.2\%$ of German Gemini generations were classifier-flagged as translated, none in the annotation set were flagged by our native annotator. For Llama, the results were more balanced: the annotator classified $37.5\%$ as translated, including $12.5\%$ marked as possibly either translated or generated (counted here as \textit{Translated}). In Spanish, results are similar: for both Gemini and Llama, $28.6\%$ of target language generations were classified as translated (compared to $56.6\%$ and $34.7\%$ by the classifier). 

We also note that most annotator rationales cited unusual words or word combinations, rather than broader issues such as abnormal frequencies of common words. Though these annotation sets are small, they suggest MLLMs may indeed be better able to avoid human-noticeable translationese errors. 

\section{Conclusion}

Through the use of state-of-the-art methods for detecting markers of translation in text, we conduct an analysis of multilingual LLM-generated text to examine its level of linguistic naturalness. We generate text in several diverse languages, creating a dataset which allows us to examine how the amount of translation interference or apparent author fluency change in different settings.

Results demonstrate that MLLMs do in fact generate text that resembles translationese, and these effects are consistently exacerbated when generating in non-English languages. By examining individual linguistic features and a set of native human annotations, however, we see that features which are more human-observable, such as unnatural or uncommon words, are more effectively avoided by the LLMs, likely due to human preference training. 

An outstanding question which remains for future work is determining the exact cause of this LLM-generated translationese: while a body of research suggests the possibility of some sort of internal translation taking place, it is also possible these features emerge as an artifact of using translated data during training. Additional experiments, such as an analysis of models with documented pretraining data compositions, could help determine whether observed translationese features are inherited from translated training data or arise as a byproduct of cross-lingual processing during inference.

\section{Limitations}

Our human annotation study is small (60 sentences per language, 2 languages) and relies on annotators personally known to the authors: native speakers with university education, but not independently recruited. We treat these results as a complement to our automated metrics, rather than a standalone finding, and future work would benefit from a larger, independently recruited pool.
Additionally, our native-speaker Pashto data is drawn from Wikipedia, and some articles may have originally been translated from English or another language before publication---a known risk for lower-resource Wikipedia editions that may partially explain our Pashto classifier's relatively lower accuracy.

\bibliography{colm2026_conference}
\bibliographystyle{colm2026_conference}

\appendix

\section{Full Data Descriptions}
\label{sec:data_app}

\subsection{Training Set}
\label{sec:trainset}

\begin{table}[H]
\centering
\small
\begin{tabular}{c|c|c|c|c|c|c}
\toprule
\multicolumn{2}{c|}{\textbf{Configuration}} 
    & \multicolumn{5}{c}{\textbf{Language}} \\
\cmidrule(lr){1-2} \cmidrule(lr){3-7}
\textbf{Generator} & \textbf{Translator} 
    & \textbf{German} & \textbf{Spanish} & \textbf{Greek} & \textbf{Pashto} & \textbf{English} \\
\midrule
\textbf{Human*}      & \textbf{N/A}      & 3200 & 3200 & 2800 & 400 & 3200 \\
\textbf{Human}      & \textbf{Human}    & 3200 & 3200 & 2800 & 400 & 3200 \\
\bottomrule
\end{tabular}
\caption{Final distribution of each language's training set, showing number of samples generated for each configuration/language pairing. \textit{Human*} refers to a native speaker of the target language.}
\label{traintable}
\end{table}

\subsection{Test Set}
\label{sec:testset}

\begin{table}[H]
\centering
\small
\begin{tabular}{c|c|c|c|c|c|c}
\toprule
\multicolumn{2}{c|}{\textbf{Configuration}} 
    & \multicolumn{5}{c}{\textbf{Language}} \\
\cmidrule(lr){1-2} \cmidrule(lr){3-7}
\textbf{Generator} & \textbf{Translator} 
    & \textbf{German} & \textbf{Spanish} & \textbf{Greek} & \textbf{Pashto} & \textbf{English} \\
\midrule
\textbf{Human*}      & \textbf{N/A}      & 800 & 800 & 700 & 100 & 800 \\
\textbf{Human}      & \textbf{Human}    & 800 & 800 & 700 & 100 & 800 \\
\textbf{Human}      & \textbf{Machine}      & 800 & 800 & 700 & 100 & 800 \\
\textbf{Human}      & \textbf{Gemini}      & 800 & 800 & 700 & 100 & 800 \\
\textbf{Human}      & \textbf{Llama}      & 800 & 800 & 700 & 100 & 800 \\
\textbf{Gemini}       & \textbf{N/A}       & 800 & 800 & 700 & 100 & 800 \\
\textbf{Gemini}        & \textbf{Machine}      & 800 & 800 & 700 & 100 & 800 \\
\textbf{Llama}       & \textbf{N/A}       & 800 & 800 & 700 & 100 & 800 \\
\textbf{Llama}        & \textbf{Machine}      & 800 & 800 & 700 & 100 & 800 \\
\bottomrule
\end{tabular}
\caption{Final distribution of each language's test set, showing number of samples generated for each configuration/language pairing. \textit{Human*} refers to a native speaker of the target language.}
\label{datatable}
\end{table}

\section{LLM Translation Results}
\label{sec:llm_trans}

\begin{table}[H]
\centering
\small
\begin{tabular}{c|c|c|c|c|c}
\toprule
\textbf{Translator} 
    & \textbf{German} & \textbf{Spanish} & \textbf{Greek} & \textbf{Pashto} & \textbf{English} \\
\midrule
\textbf{Gemini}     & 100 & 99.6 & 100 & 83.0 & 92.8 \\
\textbf{Llama}      & 100 & 95.3 & 92.2 & 78.0 & 93.1 \\
\bottomrule
\end{tabular}
\caption{Results for each language of the LLM translation settings, with the values indicating percentage of samples classified as translated by the SVM. \textit{Gemini} refers to text samples which were written by native human speakers and translated by the Gemini LLM, and \textit{Llama} refers to text samples which were written by native human speakers and translated by the Llama LLM.}
\label{lt_table}
\end{table}

\section{Europarl-UdS Experiments}
\label{sec:uds}

\subsection{Data and Classifier Specifications}
\label{sec:uds_descrip}

\subsubsection{Training Set}
\begin{table}[H]
\centering
\small
\begin{tabular}{c|c|c|c|c}
\toprule
\multicolumn{2}{c|}{\textbf{Configuration}} 
    & \multicolumn{3}{c}{\textbf{Language}} \\
\cmidrule(lr){1-2} \cmidrule(lr){3-5}
\textbf{Generator} & \textbf{Translator} 
    & \textbf{German} & \textbf{Spanish} & \textbf{English} \\
\midrule
\textbf{Human*}      & \textbf{N/A}      & 1244 & 1276 & 3200 \\
\textbf{Human}      & \textbf{Human}    & 1244 & 1276 & 3200 \\
\bottomrule
\end{tabular}
\caption{Final distribution of each language's training set for UdS experiments, showing number of samples generated for each configuration/language pairing. \textit{Human*} refers to a native speaker of the target language.}
\label{traintable2}
\end{table}

\subsubsection{Test Set}
\begin{table}[H]
\centering
\small
\begin{tabular}{c|c|c|c|c}
\toprule
\multicolumn{2}{c|}{\textbf{Configuration}} 
    & \multicolumn{3}{c}{\textbf{Language}} \\
\cmidrule(lr){1-2} \cmidrule(lr){3-5}
\textbf{Generator} & \textbf{Translator} 
    & \textbf{German} & \textbf{Spanish} & \textbf{English} \\
\midrule
\textbf{Human*}      & \textbf{N/A}      & 311 & 319 & 800 \\
\textbf{Human}      & \textbf{Human}    & 311 & 319 & 800 \\
\textbf{Human}      & \textbf{Machine}      & 311 & 319 & 0 \\
\textbf{Human}      & \textbf{Gemini}      & 311 & 319 & 0 \\
\textbf{Human}      & \textbf{Llama}      & 311 & 319 & 0 \\
\textbf{Gemini}       & \textbf{N/A}       & 311 & 319 & 800 \\
\textbf{Gemini}        & \textbf{Machine}      & 311 & 319 & 800 \\
\textbf{Llama}       & \textbf{N/A}       & 311 & 319 & 800 \\
\textbf{Llama}        & \textbf{Machine}      & 311 & 319 & 800 \\
\bottomrule
\end{tabular}
\caption{Final distribution of each language's test set for UdS experiments, showing number of samples generated for each configuration/language pairing. \textit{Human*} refers to a native speaker of the target language.}
\label{datatable2}
\end{table}

\subsubsection{Classifier Accuracy Scores}

\begin{table}[H]
\centering
\captionsetup{width=.7\linewidth}
\begin{tabular}{lc}
\toprule
\textbf{Language} & \textbf{Accuracy (\%)} \\
\midrule
German  & 96.66 \\
Spanish & 88.52 \\
English & 94.16 \\
\bottomrule
\end{tabular}
\caption{The SVM's classification accuracy by language in percentages, trained on the Europarl-UdS data.}
\label{accuracies2}
\end{table}

\subsection{Results}
\label{sec:uds_results}

\subsubsection{English}
\begin{figure}[H]
\centering
\vspace*{-1cm}
\captionsetup{width=.9\linewidth}
\includegraphics[width=0.57\columnwidth,height=0.9\columnwidth,keepaspectratio]{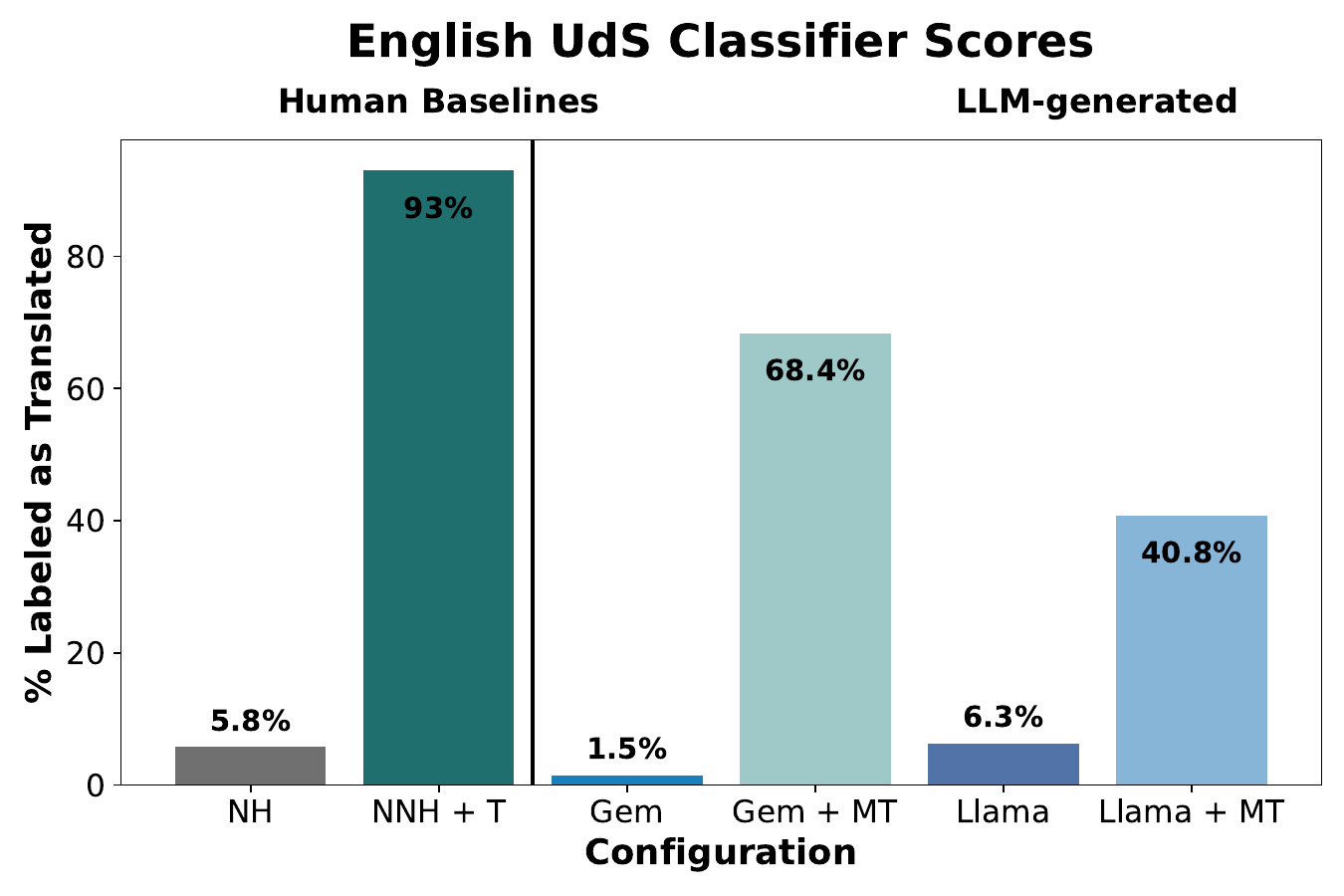}
    \caption{Full results for English on the Europarl-UdS data. See Figure \ref{english} for the complete dictionary of abbreviations.}
    \label{english_uds}
\end{figure}

\subsubsection{German}
\begin{figure}[H]
\centering
\vspace*{-1cm}
\captionsetup{width=.9\linewidth}
\includegraphics[width=0.57\columnwidth,height=0.9\columnwidth,keepaspectratio]{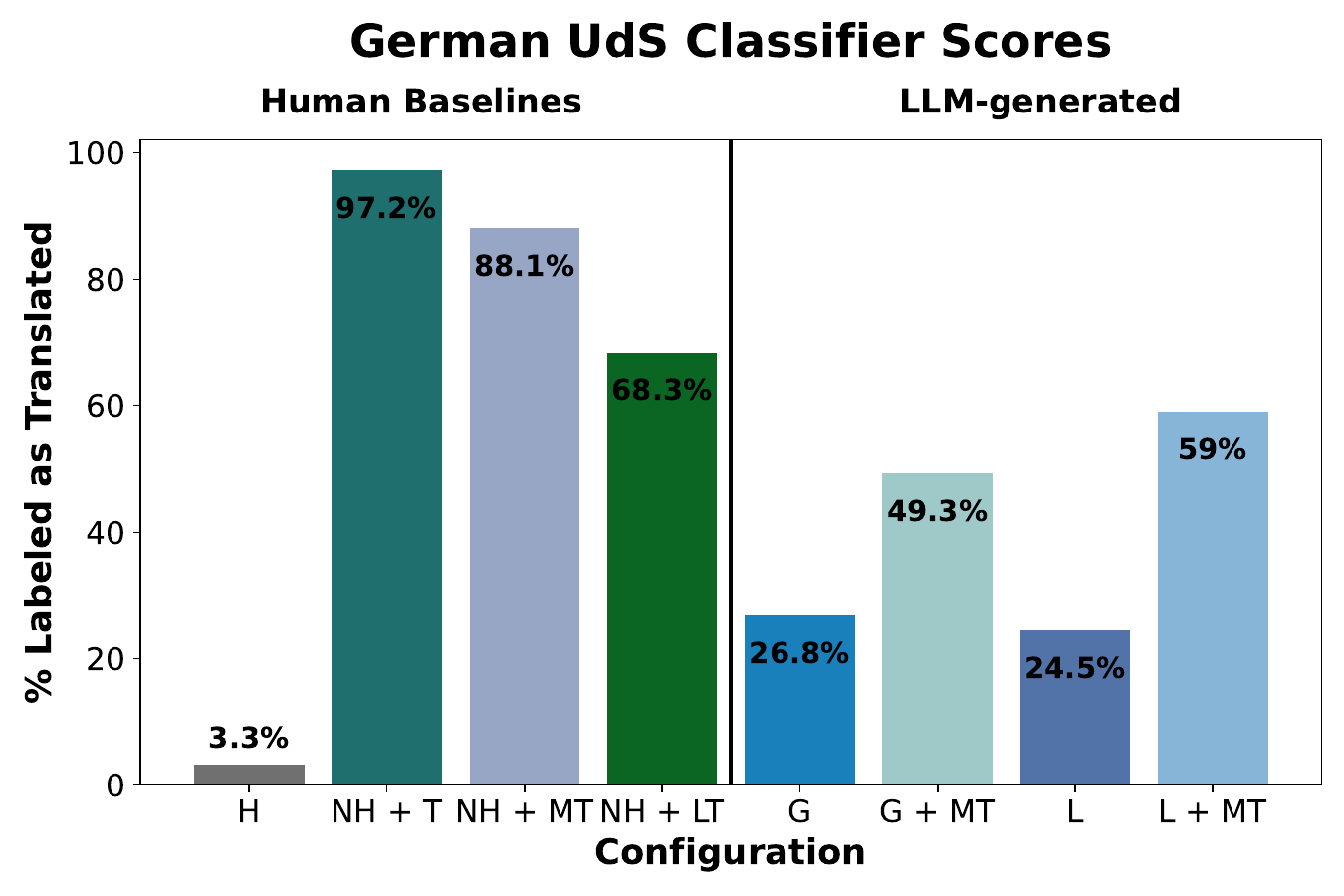}
    \caption{Full results for German on the Europarl-UdS data. See Figure \ref{english} for the complete dictionary of abbreviations.}
    \label{german_uds}
\end{figure}

\subsubsection{Spanish}
\begin{figure}[H]
\centering
\vspace*{-1cm}
\captionsetup{width=.9\linewidth}
\includegraphics[width=0.57\columnwidth,height=0.9\columnwidth,keepaspectratio]{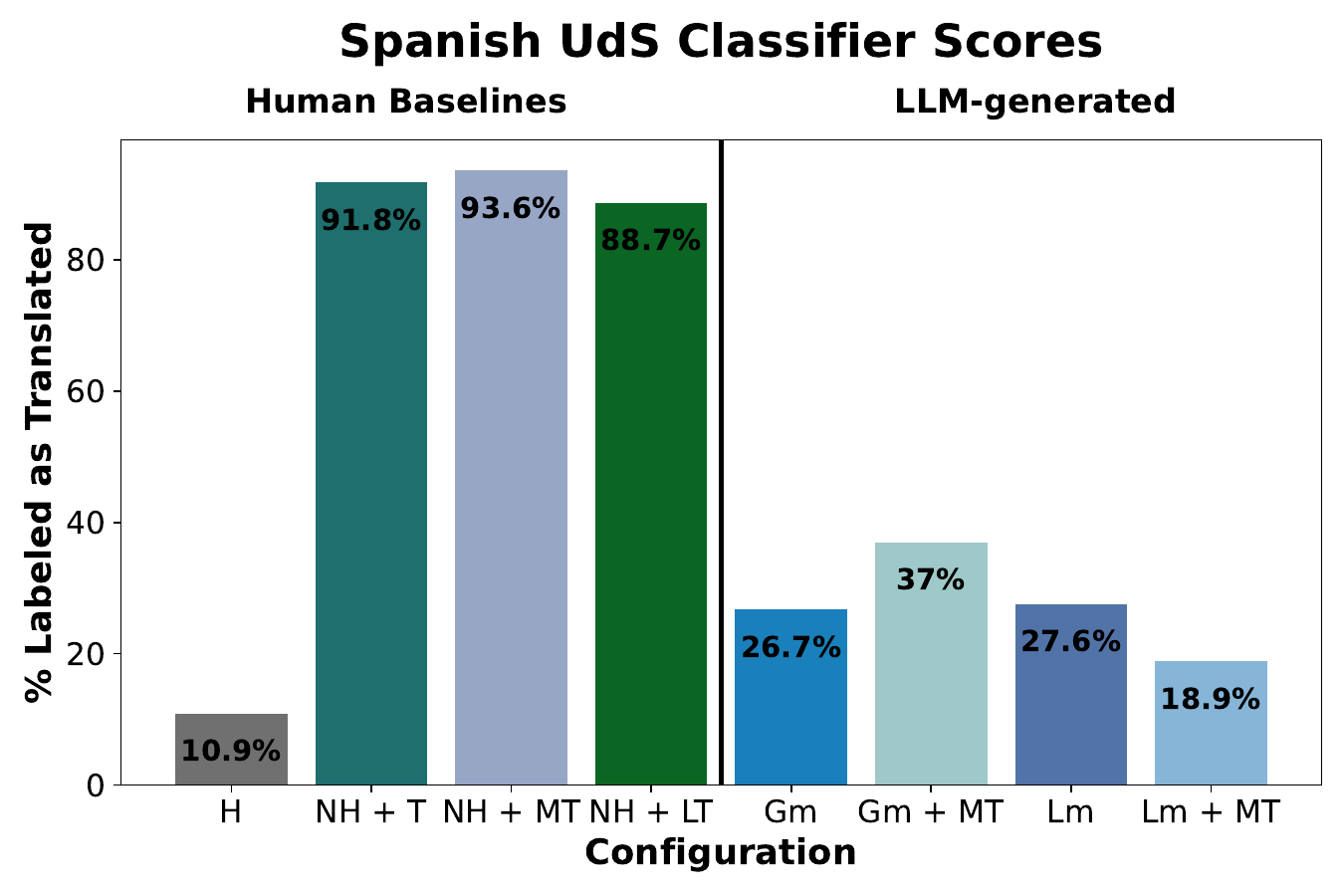}
    \caption{Full results for Spanish on the Europarl-UdS data. See Figure \ref{english} for the complete dictionary of abbreviations.}
    \label{spanish_uds}
\end{figure}

\subsection{LLM Translation Results}

\begin{table}[H]
\centering
\small
\begin{tabular}{c|c|c}
\toprule
\textbf{Translator} 
    & \textbf{German} & \textbf{Spanish}  \\
\midrule
\textbf{Gemini}     & 76.0 & 89.7  \\
\textbf{Llama}      & 60.6 & 87.8  \\
\bottomrule
\end{tabular}
\caption{Results for each language of the LLM translation settings on the UdS dataset, with the values indicating percentage of samples classified as translated by the SVM. \textit{Gemini} refers to text samples which were written by native human speakers and translated by the Gemini LLM, and \textit{Llama} refers to text samples which were written by native human speakers and translated by the Llama LLM.}
\label{lt_table}
\end{table}

\section{Full Spanish Article ANOVA Results}
\label{sec:spanish}

\subsection{El}
\begin{figure}[H]
\centering
\captionsetup{width=.9\linewidth}
\includegraphics[width=0.57\columnwidth,height=0.9\columnwidth,keepaspectratio]{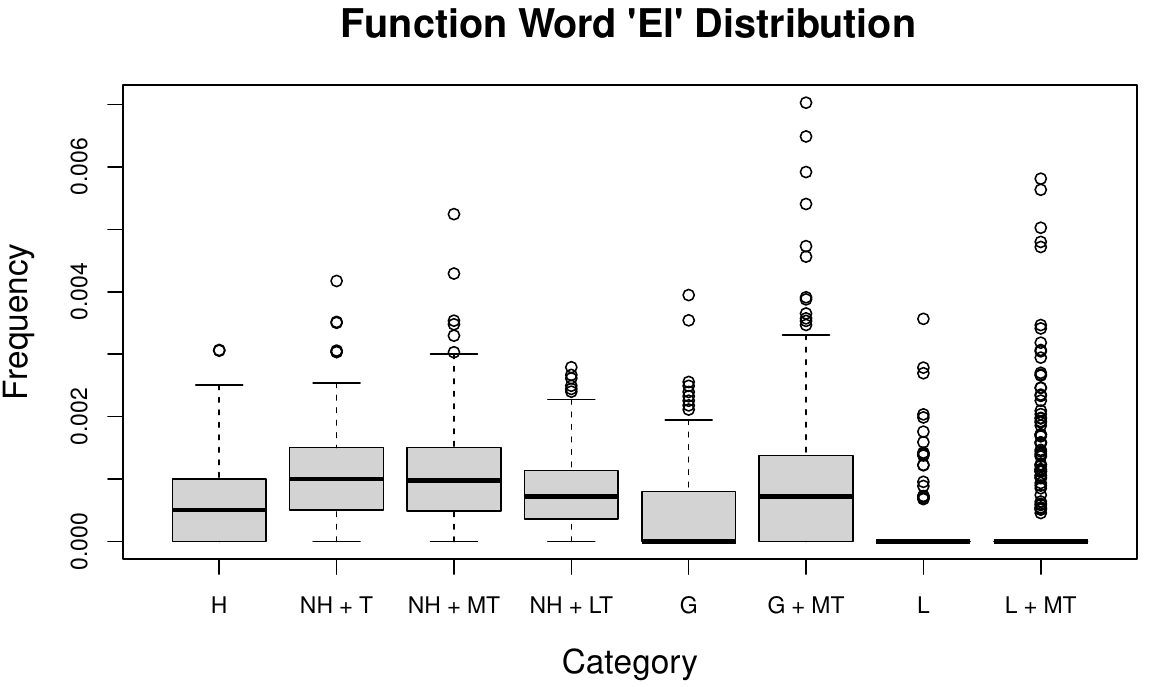}
    \caption{Distribution of the Spanish article \textit{el} across configurations. See Figure \ref{english} for the complete dictionary of abbreviations.}
    \label{fw_el}
\end{figure}

\subsection{La}
\begin{figure}[H]
\centering
\captionsetup{width=.9\linewidth}
\includegraphics[width=0.57\columnwidth,height=0.9\columnwidth,keepaspectratio]{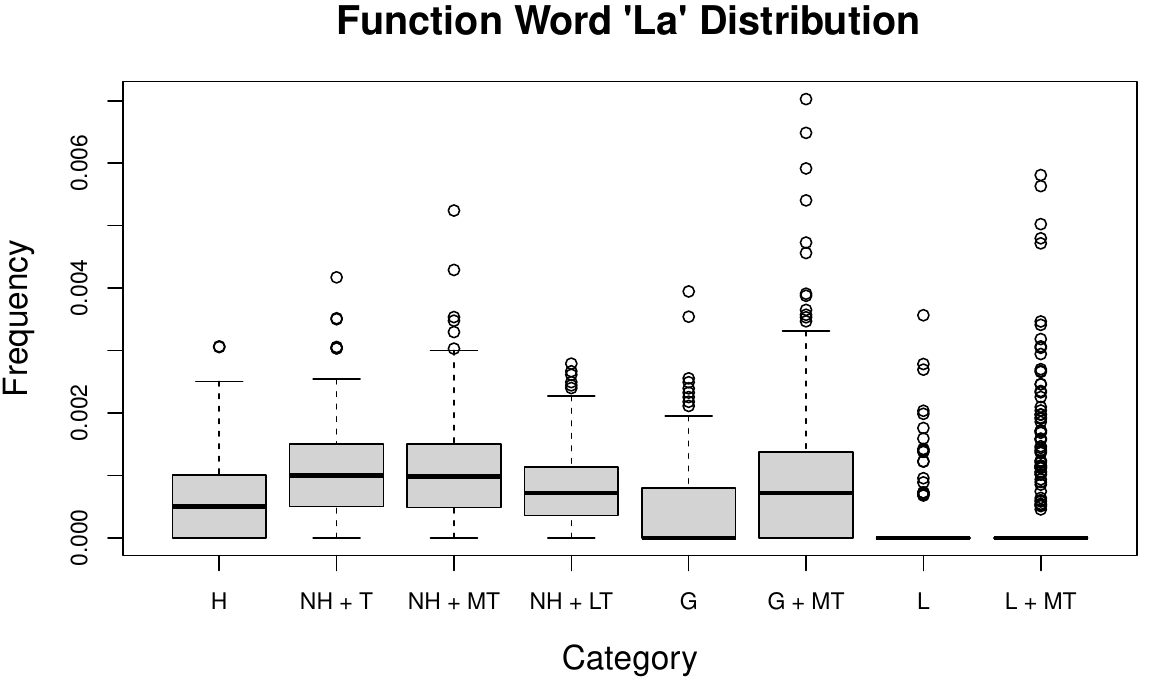}
    \caption{Distribution of the Spanish article \textit{la} across configurations. See Figure \ref{english} for the complete dictionary of abbreviations.}
    \label{fw_la}
\end{figure}

\subsection{Las}
\begin{figure}[H]
\centering
\captionsetup{width=.9\linewidth}
\includegraphics[width=0.57\columnwidth,height=0.9\columnwidth,keepaspectratio]{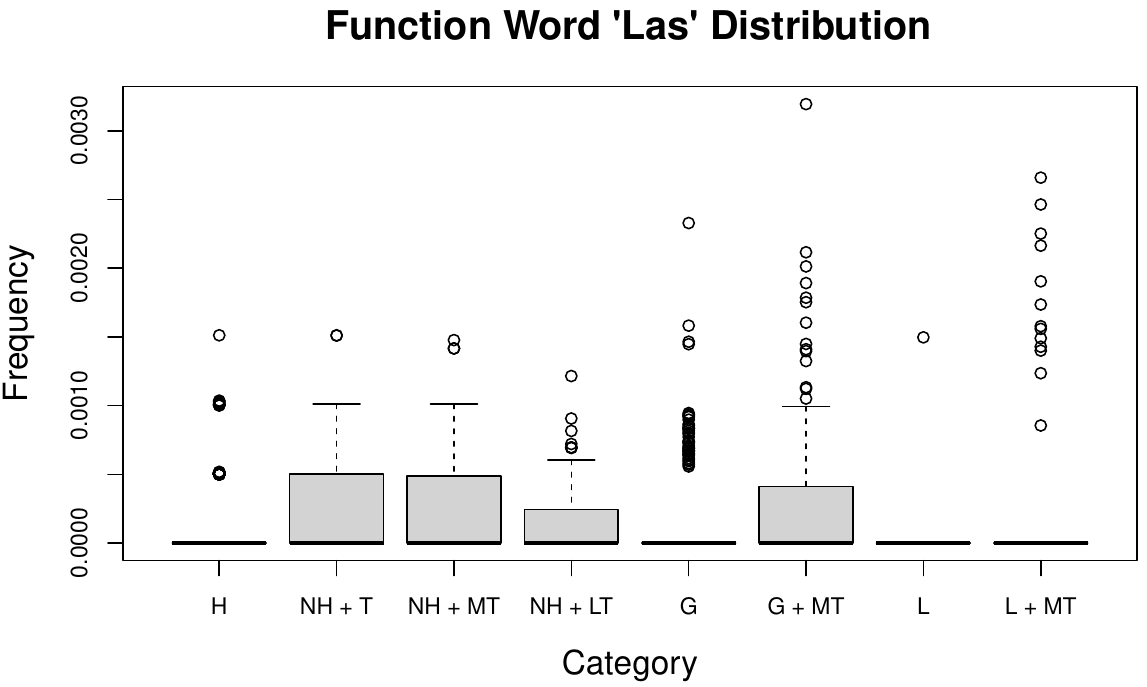}
    \caption{Distribution of the Spanish article \textit{las} across configurations. See Figure \ref{english} for the complete dictionary of abbreviations.}
    \label{fw_las}
\end{figure}

\subsection{Los}
\begin{figure}[H]
\centering
\captionsetup{width=.9\linewidth}
\includegraphics[width=0.57\columnwidth,height=0.9\columnwidth,keepaspectratio]{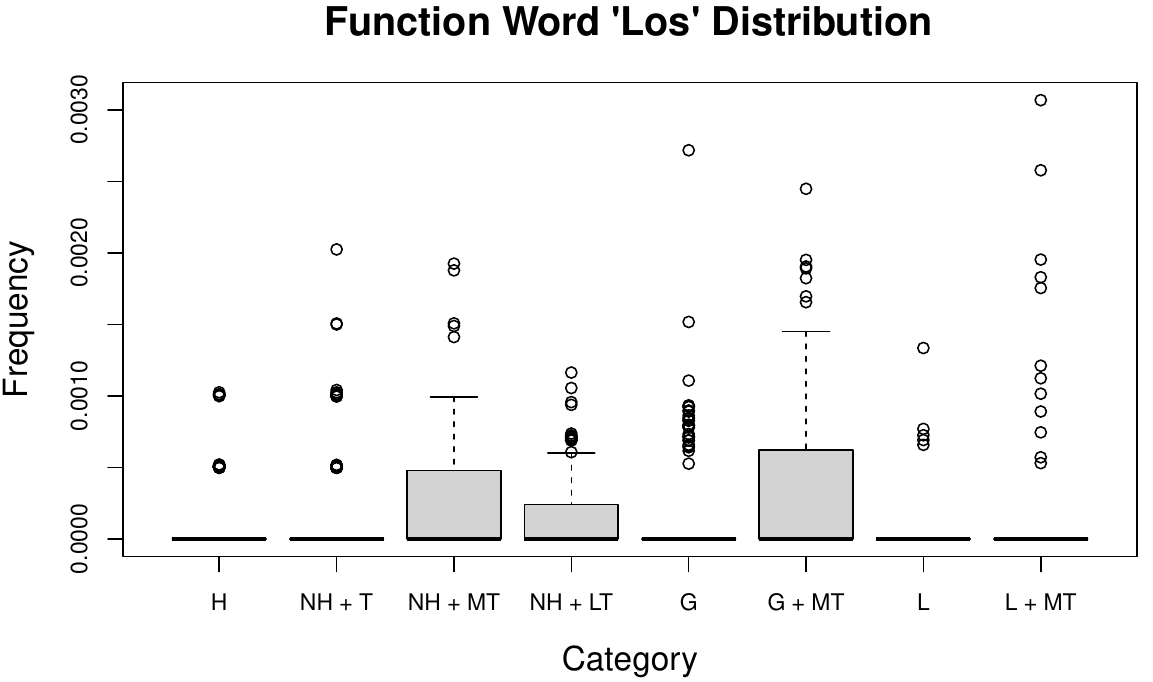}
    \caption{Distribution of the Spanish article \textit{los} across configurations. See Figure \ref{english} for the complete dictionary of abbreviations.}
    \label{fw_los}
\end{figure}

\end{document}